\documentclass{article}
\usepackage{log_2024}						

\usepackage{booktabs}						
\usepackage{multirow}						
\usepackage{amsfonts}						
\usepackage{graphicx}						
\usepackage{duckuments}						

\usepackage[numbers,compress,sort]{natbib}	

\usepackage{amssymb,amsmath,amsthm}

\usepackage{amsmath,amsfonts,bm}
\usepackage{amsthm}

\theoremstyle{plain}
\newtheorem{theorem}{Theorem}[section]

\theoremstyle{definition}

\theoremstyle{remark}

\def\eqref#1{equation~\ref{#1}}

\def\1{\bm{1}}

\DeclareMathAlphabet{\mathsfit}{\encodingdefault}{\sfdefault}{m}{sl}
\SetMathAlphabet{\mathsfit}{bold}{\encodingdefault}{\sfdefault}{bx}{n}

\usepackage{subfigure}
\usepackage{subcaption}
\usepackage{caption}
\usepackage{algorithm}
\usepackage{algorithmic}
\usepackage{url}
\usepackage{adjustbox}
\usepackage{xcolor}
\usepackage{colortbl}

\definecolor{Gray}{gray}{0.85}
\newcolumntype{a}{>{\columncolor{Gray}}c}

\title[Enhancing Topological Dependencies in Spatio-Temporal Graphs with Cycle Message Passing Blocks]{Enhancing Topological Dependencies in Spatio-Temporal Graphs with Cycle Message Passing Blocks}

\author[M. Lee et al.]{%
Minho Lee\thanks{Equal contribution. $\dagger$ Correspondence to: Yun Young Choi <young@aiae.co>}\\
AI Aided Engineering\\
\email{minho.lee@aiae.co}\And
Yun Young Choi$\dagger$\footnotemark[1]\\
AI Aided Engineering\\
\email{young@aiae.co}\And
Sun Woo Park\\
Max Planck Institute for Mathematics\\
\email{s.park@mpim-bonn.mpg.de}\And
Seunghwan Lee\\
AI Aided Engineering\\
\email{seunghwan.lee@aiae.co}\And
Joohwan Ko\\
University of Massachusetts Amherst\\
\email{joohwanko@cs.umass.edu}\AND
Jaeyoung Hong\\
AI Aided Engineering\\
\email{jaeyoung.hong@aiae.co}
}

\begin{document}

\maketitle

\begin{abstract}
Graph Neural Networks (GNNs) and Transformer-based models have been increasingly adopted to learn the complex vector representations of spatio-temporal graphs, capturing intricate spatio-temporal dependencies crucial for applications such as traffic datasets. Although many existing methods utilize multi-head attention mechanisms and message-passing neural networks (MPNNs) to capture both spatial and temporal relations, these approaches encode temporal and spatial relations independently, and reflect the graph's topological characteristics in a limited manner.
In this work, we introduce the Cycle to Mixer (Cy2Mixer), a novel spatio-temporal GNN based on topological non-trivial invariants of spatio-temporal graphs with gated multi-layer perceptrons (gMLP). The Cy2Mixer is composed of three blocks based on MLPs: A temporal block for capturing temporal properties, a message-passing block for encapsulating spatial information, and a cycle message-passing block for enriching topological information through cyclic subgraphs.
We bolster the effectiveness of Cy2Mixer with mathematical evidence emphasizing that our cycle message-passing block is capable of offering differentiated information to the deep learning model compared to the message-passing block. Furthermore, empirical evaluations substantiate the efficacy of the Cy2Mixer, demonstrating state-of-the-art performances across various spatio-temporal benchmark datasets. The source code is available at \url{https://github.com/leemingo/cy2mixer}.
\end{abstract}

\section{Introduction}
Spatio-temporal forecasting predicts future events or states involving both spatial and temporal components, with traffic forecasting being one of the most representative examples among these problems. Traffic data is frequently conceptualized as a spatio-temporal graph, where road connections and traffic flows gleaned from sensors are represented as edges and nodes~\citep{diao2019dynamic, wang2020traffic}, and traffic forecasting aims to predict future traffic in road networks based on preceding traffic data~\citep{li2017diffusion, deng2021st, shao2022spatial, shao2022decoupled, zhu2023correlation}. 
Given this, many turn to Graph Neural Networks (GNNs), especially those using Message Passing Neural Networks (MPNNs), to study this data. 
GNNs based on MPNNs have been increasingly adopted to learn the complex vector representations of spatio-temporal graphs, capturing intricate dependencies in traffic datasets~\citep{yu2017spatio, li2021spatialtemporal,choi2022graph}.
However, traditional message passing-based methodologies have exhibited limitations,  including the inability to adequately account for temporal variations in traffic data and over-smoothing problems~\citep{jiang2023pdformer}. Following the success of Transformers in various domains, they have been employed either independently or synergetically with MPNNs to address these challenges in traffic forecasting~\citep{zheng2020gman, xu2020spatial, jiang2023pdformer}, some implementations of which are coupled with preprocessing techniques like Dynamic Time Warping (DTW)~\citep{berndt1994using}, showcasing competitive performance outcomes~\citep{jiang2023pdformer}.

Previous research efforts have predominantly focused on employing additional algorithms or incorporating intricate structures to capture the complex patterns inherent in spatio-temporal graphs. This often results in heuristics that inflate the complexity of structure and computational costs. Additionally, many of these studies have leaned heavily on experimental results to justify their performance improvements, lacking comprehensive explanations or theoretical grounding. To address these challenges, we start with a mathematical hypothesis that leverages the topological non-trivial invariants of a spatio-temporal graph to enhance the predictive performances of GNNs. This hypothesis aims to discern the spatio-temporal relationships among nodes within the graph. We demonstrate that topological features shed light on facets of temporal traffic data that might be overlooked when focusing solely on the pre-existing edges in a traffic network.

\begin{figure*}[t]
    \centering
    \includegraphics[width=1\textwidth]{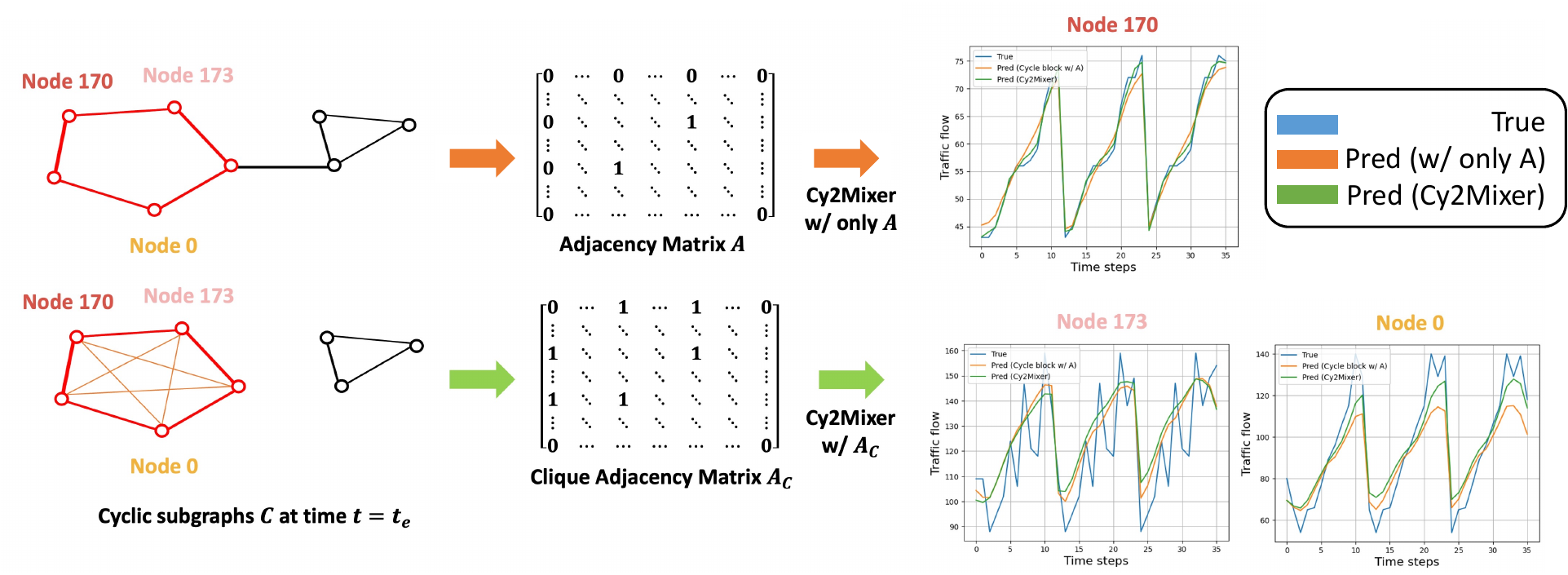}
    \caption{Prediction example of Cy2Mixer in the PEMS04 dataset. While Cy2Mixer shows similar performance between Node 170 and Node 173, which are connected via the adjacency matrix \( A \), Cy2Mixer exhibits superior performance at Node 0 when utilizing the clique adjacency matrix \( A_C \), indicating the effectiveness of the cycle block in capturing cyclic subgraph relationships.} 
    \label{Fig:pred_ex}
\end{figure*}

Building on this mathematical foundation, we propose a simple yet unique model, Cycle to Mixer (Cy2Mixer), that integrates gated Multi-Layer Perceptron (gMLP)~\cite{liu2021pay} and MPNNs supplemented with topological non-trivial invariants in graphs, to enhance predictive accuracy. Our Cy2Mixer layer comprises of three key components: (a) Temporal block: Seizes the temporal characteristics of all nodes. (b) Spatial message-passing block: Encompasses spatial relationships along with local neighborhoods. (c) Cycle message-passing block: Captures supplementary information among nodes in cyclic subgraphs. 
Notably, our cycle message-passing block aims to encapsulate topological features, enabling a finer comprehension of the intricate connectivity patterns within the network inspired by Cy2C-GNNs~\citep{choi2022cycle, choi2024topology}. We validate our methodology using traffic datasets, showcasing the competitive edge of our proposed models. Furthermore, a qualitative assessment elucidates the efficacy of topological invariants in enhancing the Cy2Mixer's predictive performance as demonstrated in Figure \ref{Fig:pred_ex}, showing improved accuracy for nodes connected through cyclic subgraphs when using the clique adjacency matrix. While our experiments focus on traffic datasets as representative spatio-temporal datasets, we also conducted experiments to demonstrate that our methodology performs well on other spatio-temporal datasets. Our work demonstrates that our approach is not limited to a single task but performs effectively across various tasks, while also evaluating the efficiency of preprocessing time.

The contributions of this work are as follows: 
(1) We use homotopy invariance of fundamental groups and homology groups to deduce that topological non-trivial invariants of a spatio-temporal network become a contributing factor for influencing the spatio-temporal forecast. 
(2) We propose a simple yet novel network, Cy2Mixer, based on the gMLP and Cy2C-GNNs inspired from the theory of covering spaces. (3) We show that the proposed model not only demonstrates efficiency in preprocessing time but also consistently delivers superior performance across a variety of spatio-temporal datasets.

\section{Related Works}
In recent years, deep learning approaches have emerged as powerful tools in the spatio-temporal modeling by leveraging their ability to automatically learn complex, nonlinear spatio-temporal patterns, and traffic forecasting is one of the primary applications of spatio-temporal modeling. Convolutional Neural Networks (CNNs) have been effectively applied to traffic forecasting to capture spatio-temporal patterns~\cite{zhang2017deep, li2017diffusion, lin2020preserving}. With the success of Transformers in fields like natural language processing and computer vision, owing to their ability to model complex relationships over long sequences without relying on recurrent structures, they have been effectively adapted to traffic forecasting by leveraging self-attention mechanisms to capture long-range spatio-temporal dependencies~\cite{xu2020spatial, jiang2023pdformer, liu2023spatio}. Meanwhile, to capture the dynamic spatial correlations within traffic networks, Graph Neural Networks (GNNs) have become a prominent approach for traffic forecasting due to their ability to model complex dependencies~\cite{yu2017spatio, pan2019urban, wu2019graph, choi2022graph}, and some studies have integrated self-attention mechanisms with GNNs to further enhance the modeling of spatial relationships within traffic networks~\cite{zheng2020gman, guo2021learning}. Some recent studies adapt traffic graph structures through learnable adjacency matrices and probabilistic modeling~\cite{wu2020connecting, shang2021discrete} and integrate hierarchical topological views to capture multi-level spatial patterns~\cite{wu2020learning, ma2023rethinking}. However, many existing approaches that incorporate spatial information in traffic networks have relied on either limited methods or overly heuristic strategies with auxilary information from road networks, which may be challenging to apply to other spatio-temporal tasks. These methods often struggle to comprehensively capture the full complexity of spatial dependencies, leading to suboptimal modeling of traffic dynamics.

\section{Preliminaries}
In this section, we provide explanations for some notations and define the traffic prediction problem.
\subsection{Message passing neural networks (MPNNs)}
\textcolor{black}{Each $m$-th layer $H^{(m)}$ of MPNNs and hidden node attributes $h_v^{(m)}$ with dimension $k_m$ are given by:}
\begin{align}
    \begin{cases}
        h_v^{(m)} := \text{COMBINE}^{(m)} \left( h_v^{(m-1)}, h_u^\prime \right)\\
        h_v^{(0)} := X_v
    \end{cases} \\
    \text{where }h_u^\prime = \text{AGGREGATE}_v^{(m)}\left( \left\{\!\!\left\{ h_u^{(m-1)} \; | \; u \in N(v) \right\}\!\!\right\} \right).
\end{align}
$X_v$ is the initial node attribute at $v$ and {$N(v)$ is the set of neighborhood nodes}. Note that $\text{AGGREGATE}_v^{(m)}$ is a function that aggregates features of nodes adjacent to $v$, and $\text{COMBINE}^{(m)}$ is a function which combines features of the node $v$ with those of nodes adjacent to $v$.  

\subsection{gated Multi-Layer Perceptron (gMLP)} 
gMLP has been shown to achieve comparable performance to Transformer models across diverse domains, including computer vision and natural language processing, with improved efficiency~\citep{liu2021pay}. \textcolor{black}{With the given input $X \in \mathbb{R}^{n \times d}$, where $n$ denotes sequence length and $d$ denotes dimension, it can be defined as:}
\begin{align}
    Z=\sigma(X U), \quad \tilde{Z}=s(Z), \quad Y=\tilde{Z} V,
\end{align}
where $\sigma$ represents the activation function, which is GeLU~\citep{hendrycks2023gaussian} in this context, and $U$ and $V$ correspond to linear projections based on the channel (feature) dimensions, serving roles similar to the feed-forward network in Transformer. The layer $s(\cdot)$ corresponds to the Spatial Gating Unit, which is responsible for capturing cross-token interactions.
We construct the aforementioned layer as:
\begin{align}
    s(Z) = Z_{1} \odot f_{W,b}(Z_{2}),
\end{align}
\textcolor{black}{where $f_{W,b}$ denotes a linear projection $f_{W,b}(Z) = WZ+b$. Here, $W$ and $b$ refer to the weight matrix and bias, respectively. $Z_{1}$ and $Z_{2}$ denote two independent components split from $Z$ along the channel dimension, and $\odot$ denotes element-wise multiplication.}

\subsection{Traffic Prediction Problem}
\paragraph{Traffic sensor} Traffic sensors are deployed within the traffic system to record essential information, such as the flow of vehicles on roads and the speeds of these vehicles.

\paragraph{Traffic network} Traffic network can be represented as $\mathcal{G}=(\mathcal{V}, \mathcal{E}, A, A_C)$, where $\mathcal V = \{v_1, \cdots, v_N\}$ denotes the set of $N$ nodes representing sensors within the traffic network ($|V|=N$).
Next,  $\mathcal{E} \subseteq \mathcal{V} \times \mathcal{V}$ represents the set of edges, and the adjacency matrix  $A$ of the network $\mathcal{G}$ can be obtained based on the distances between nodes.
Additionally, $A_C$ is the clique adjacency matrix of the network, which is incorporated into the GNN architecture. 
The detailed procedure for obtaining the clique adjacency matrix is provided in Appendix A.7.
The matrices $A \in \mathbb{R}^{N \times N}$ and $A_C \in \mathbb{R}^{N \times N}$ are time-independent input variables since the structure remains unchanged over time. As will be noted in Section 4 and Appendix A.6, cyclic substructures of traffic network are crucial in understanding additional conditions imposed in accurately predicting traffic signals. As demonstrated in previous studies~\citep{choi2022cycle}, utilizing clique adjacency matrices $A_C$ can effectively boost capabilities of conventional deep learning architectures in incorporating cyclic substructures of traffic networks. As such, we use the clique adjacency matrix $A_C$ as additional input to our traffic network data.

\paragraph{Traffic signal} The traffic signal $X_t \in \mathbb{R}^{N \times C}$ represents the data measured at time $t$ across $N$ nodes in the network.
In this context, $C$ denotes the number of features being recorded by the sensors, and in this study, it represents the flow of the road network.

\paragraph{Problem formalization} Our objective is to train a mapping function $f$ to predict future traffic signals by utilizing the data observed in the previous $T$ steps, which can be illustrated as follows:
\begin{align}
\left[X_{(t-T+1)}, \cdots, X_t ; \mathcal{G}\right] \stackrel{f}{\longrightarrow}\left[X_{(t+1)}, \cdots, X_{\left(t+T^{\prime}\right)}\right].
\end{align}

\section{Mathematical Backgrounds}

One of the prominent ways to mathematically quantify the effectiveness and discerning capabilities of GNNs is to interpret the graph dataset as a collection of 1-dimensional topological spaces $\mathcal{G} := \{G_i\}_{i}$, and identify the given GNNs as a function $GNN: \mathcal{G} \to \mathbb{R}^k$ which represents graphs in a given dataset as real $k$-dimensional vectors. These approaches were carefully executed in previous literature, which focused on pinpointing the discernability of various architectural designs for improving GNNs, as seen in \cite{choi2022cycle, bodnar21cw, horn2021topological, park2022pwlr, bernardez2024icml}. Throughout these references, the characterizations of topological invariants of such 1-dimensional topological spaces provided grounds for verifying whether such GNNs can effectively capture geometric non-trivial properties of graph datasets, such as cyclic substructures or connected components of graphs. In light of these previous studies, it is hence of paramount interest to reinterpret temporal graph datasets as a collection of higher dimensional topological spaces and understand what non-trivial properties of such topological spaces the novel deep learning techniques processing temporal graph datasets should encapsulate.

\begin{figure*}[t]
    \centering
    \includegraphics[width=0.78\textwidth]{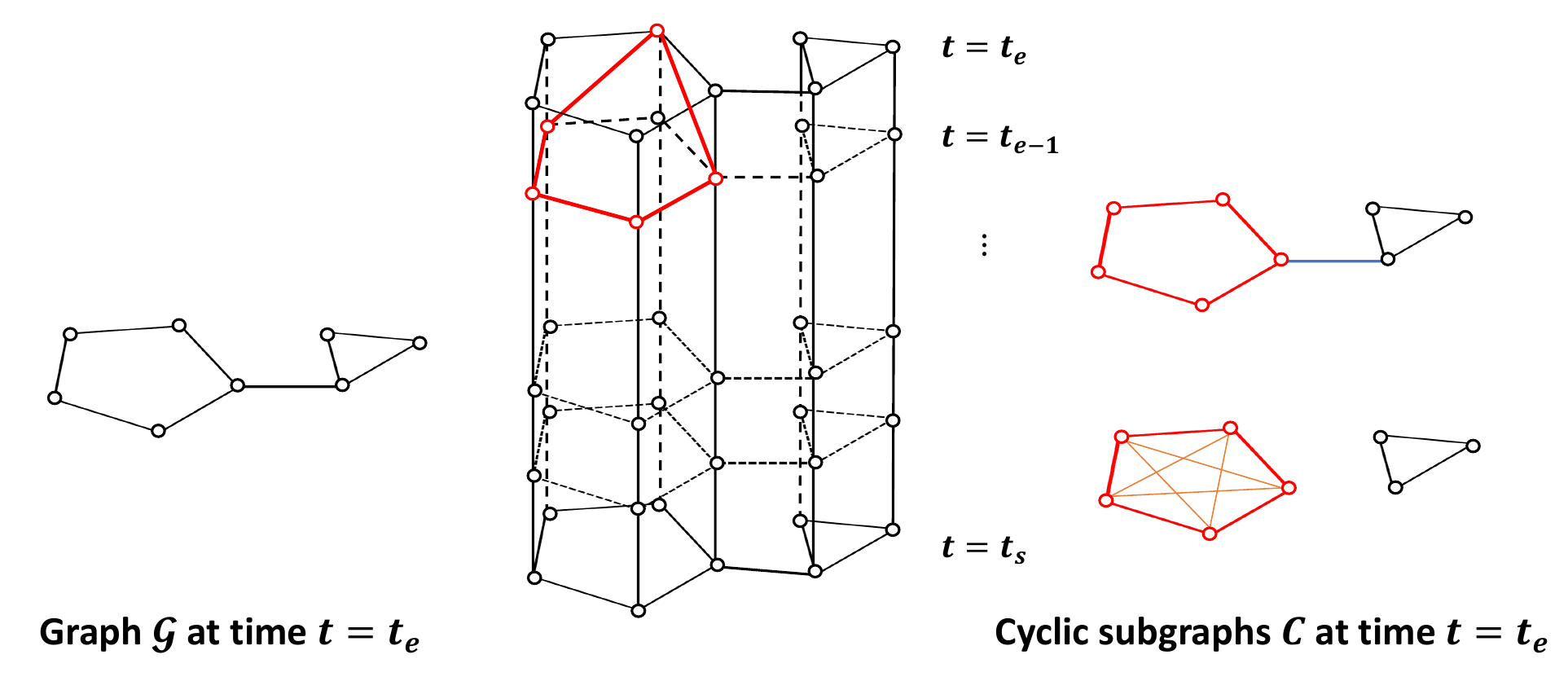}
    \caption{An illustration of lifting a cyclic subgraph of the traffic network $\mathcal{G}$ to a temporal cyclic subgraph of the topological space $\mathcal{G} \times I$ representing the traffic dataset. Every temporal cyclic subgraph of the traffic dataset $\mathcal{G} \times I$ can be obtained from cyclic subgraph of the underlying traffic network $\mathcal{G}$ by using Theorem \ref{theorem:cycle_temporal}. By transforming cyclic subgraphs into cliques by adding and deleting edges suitably, Cy2Mixer effectively models how traffic signals measured in distinct nodes and time can affect each other.}
    \label{Fig:background}
\end{figure*}

\paragraph{Topological Space} 
We interpret the traffic dataset as a 2-dimensional topological space constructed from the traffic network $\mathcal{G}$, and traffic signals as a function defined over the 2-dimensional topological space. To elaborate, the previous section indicates that we can identify a traffic network with a graph $\mathcal{G} = (\mathcal{V}, \mathcal{E}, A, A_C)$, and a traffic signal at time $t$, denoted by $X_t \in \mathbb{R}^{N \times C}$, as a collection of functions $\{f_t: \mathcal{G} \to \mathbb{R}^{N \times C}\}_{t}$, each of which represents measurements taken over the graph $\mathcal{G}$ at time $t$. Notice that the continuous time variable $t$ parametrizes a closed interval $I := [t_s, t_e]$, where $t_s$ and $t_e$ denote the start and end time of the traffic dataset. This observation allows us to identify a temporal traffic network as a topological space $\mathcal{G} \times I$, and the collection of traffic signals varying with respect to a time variable $t$ as a function over $\mathcal{G} \times I$ satisfying the following condition:
\begin{align}
\begin{split}
    X :& \mathcal{G} \times I \to \mathbb{R}^{N \times C}\\
    X(g, t) &= X_t(g) \text{ for all } t \in I.
\end{split}
\end{align}

\paragraph{Temporal cyclic structures}  One of the prominent topological properties of a graph $\mathcal{G}$ is its cyclic subgraphs. Cyclic substructures of traffic networks are crucial in determining additional obstructions imposed on forecasting traffic signals. To see this, the existence of such cycles indicates that there exists a pair of nodes $v,w \in \mathcal{V}$ which are connected by at least two distinct paths, obtained from moving along the edges of the graph. In terms of traffic dataset, the existence of a cyclic subgraph indicates that there are at least two paths that a flow of traffic can move along from one point to another. Hence, it is crucial to understand how cyclic subgraphs of a given traffic network $\mathcal{G}$ affect future traffic signals. This can be achieved by understanding possible cyclic substructures inherent in a traffic signal $X: \mathcal{G} \times I \to \mathbb{R}^{N \times C}$. The following result shows that cyclic substructures of a traffic signal can be fully understood from the cycle bases of a traffic network $\mathcal{G}$. We leave the proof of the theorem as well as remarks on mathematical relations between cyclic subgraphs and temporal traffic data in Appendix A.6.

\begin{theorem}
\label{theorem:cycle_temporal}
    Given any choice of $t_0 \in I$, let $\pi_{t_0}: \mathcal{G} \times I \to \mathcal{G} \times \{t_0\} \cong \mathcal{G}$ be the projection map which sends the interval $I$ to a singleton set $\{t_0\}$. Let $\mathcal{C}_{\mathcal{G} \times I} := \{C_1, C_2, \cdots, C_n\}$ be a cycle basis of the topological space $\mathcal{G} \times I$. Then the set
    \begin{equation}
        \pi_{t_0} \left( \mathcal{C}_{\mathcal{G} \times I} \right) := \{\pi_{t_0}(C_1), \cdots, \pi_{t_0}(C_n)\}
    \end{equation}
    is a cycle basis of the traffic network $\mathcal{G}$.
\end{theorem}

\paragraph{Temporal cliques} Theorem \ref{theorem:cycle_temporal} demonstrates that every cyclic structure of temporal graph datasets $\mathcal{G} \times I$, originating from lifting the cyclic structures of the baseline traffic network $\mathcal{G}$ to any time period, can be captured using the projection map $\pi_{t_0}$. In fact, the lifting procedure does not require that the cyclic structure over $\mathcal{G} \times I$ has to be defined over a fixed time $t$. It can be lifted to a cyclic subgraph of $\mathcal{G} \times I$ such that its nodes represent traffic signals measured at different moments of time.

One of the topological invariants that conventional GNNs fail to incorporate effectively is the cyclic substructures of graphs \cite{xu2019powerful, bodnar21cw}. This is because GNNs can only distinguish any pairs of two graphs $G$ and $H$ up to isomorphism of their universal covers or unfolding trees. To overcome this issue, one can substitute cyclic subgraphs with cliques. An approach inspired by the theory of covering spaces, the strategy alters the geometry of universal covers to allow GNNs to encapsulate cyclic structures effectively~\citep{choi2022cycle}. We implement the application of the analogous operation to temporal traffic data by utilizing the clique adjacency matrix $A_C$.
The clique adjacency matrix $A_C$ allows room for GNNs to determine whether additional interactions among nodes of $\mathcal{G}$, whose measurements are taken at varying moments of time, are relevant components for forecasting traffic signals. This allows us to capture potential effects of topological invariants of traffic networks on temporal traffic measurements, as suggested from Theorem \ref{theorem:cycle_temporal}. Figure~\ref{Fig:background} illustrates how lifting a cyclic subgraph of $\mathcal{G}$ to a new cyclic subgraph of $\mathcal{G} \times I$ of varying temporal instances can enrich our prediction of future traffic signals. The two cycles are subgraphs colored in red. The temporal cyclic subgraph spans over time $t = t_0$ and the end time $t = t_e$ of the dataset. By changing the colored cyclic subgraph into cliques, we add edges to nodes on the cyclic subgraph lying in two different instances of time. These edges establish relations among measurements taken at different nodes and time periods.

\begin{figure*}[t]
    \centering
    \includegraphics[width=1\textwidth]{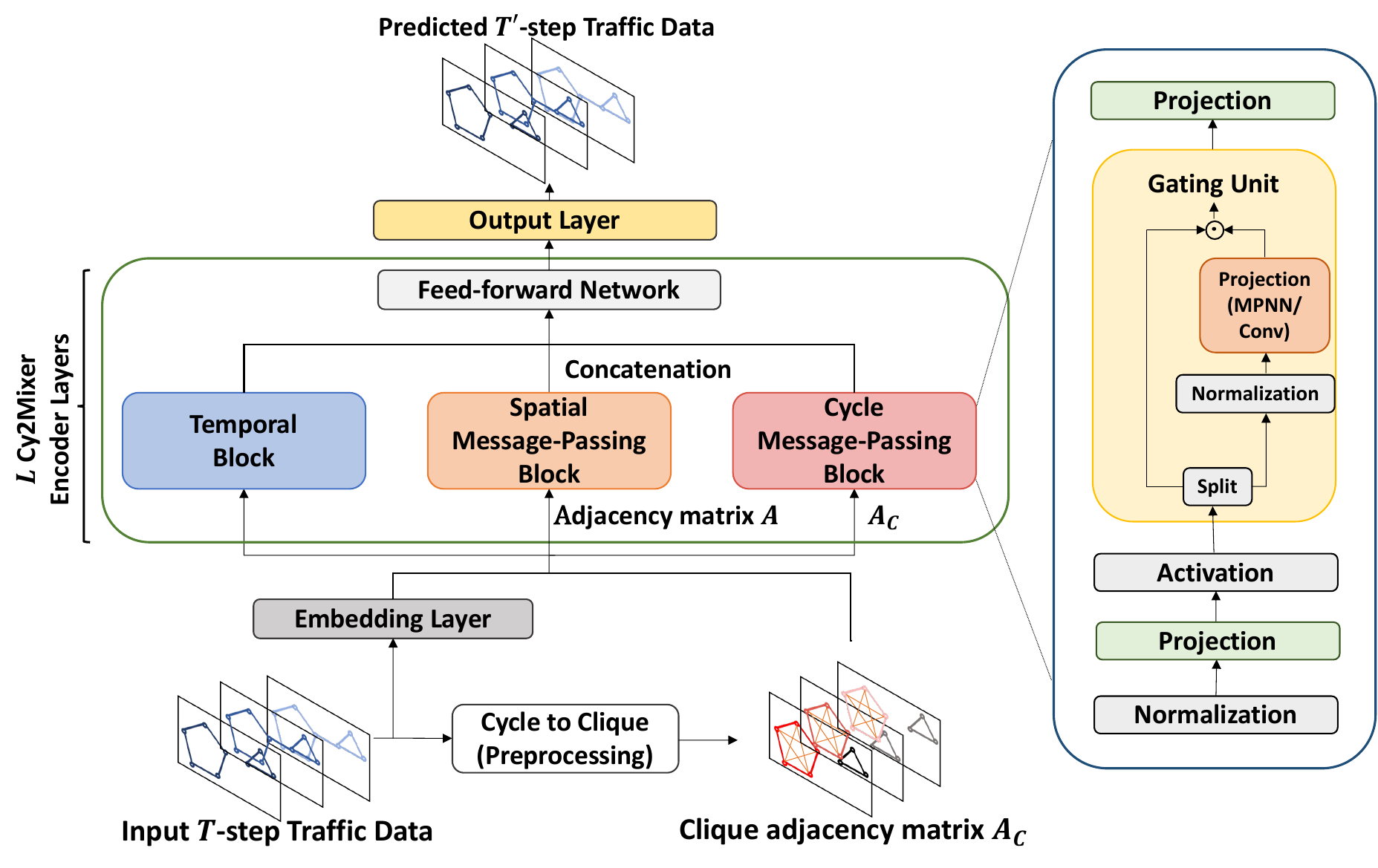}
    \caption{The overall framework of Cy2Mixer. Layers within the green box indicate the Cy2Mixer encoder layer, which comprises a temporal block, spatial message-passing block, and cycle message-passing block to ensure a comprehensive understanding of both temporal and spatial aspects.} 
    \label{Fig:framework}
\end{figure*}

\section{Methodology}
In this section, we present the architecture of Cy2Mixer and elaborate on how it distinguishes itself from other models. Figure \ref{Fig:framework} provides a comprehensive visualization of our model's framework. This architecture is primarily composed of three main components: the embedding layer, a stack of $L$ Cy2Mixer encoder layers, and the output layer. 

\subsection{Embedding Layer}
Our model follows the structure of the embedding layer from \citet{liu2023spatio}. Given the time-series data $X_{t-T+1:t}$, the feature embedding is defined as \(X_{\textit{feat}}= \text{FC}(X_{t-T+1:t}) \), where FC represents a fully connected layer and \(X_{\textit{feat}} \in \mathbb{R}^{T \times N \times d_f}\). Additionally, to account for both weekly and daily periodicities, the temporal embedding \(X_{\textit{temp}} \in \mathbb{R}^{T\times N \times d_{t}}\) is produced. This is achieved by referencing the day-of-week and timestamp-of-day data to extract two distinct embeddings, which are subsequently concatenated and broadcasted to generate the final temporal embedding with dimension $d_{t}$. 
Furthermore, to address diverse temporal patterns specific to each node, a learnable parameter is introduced in the form of an adaptive embedding of dimension $d_a$, represented as $X_{\textit{ast}} \in {\mathbb{R}^{T \times N \times d_{a}}}$, to capture and reflect latent, unobserved information within the temporal and spatial patterns of each node. These three embeddings are concatenated, forming the output $H\in \mathbb{R}^{T \times N \times d_h}$ of the embedding layer, where the dimension $d_h$ is equal to $d_f + d_t + d_a$.
\begin{align}
    H = X_{\textit{feat}}||X_{\textit{temp}}||X_{\textit{ast}}.
\end{align}
    
\subsection{Cy2Mixer Encoder Layer} 

The core architecture of the Cy2Mixer encoder layer consists of three distinct blocks, each of which calculates the projection values of \( Z \) based on the output \( H \) from the embedding layer. In this context, \( Z \) corresponds to the output of the projection layer applied to \( H \), analogous to how \( X \) in Section 2.2 undergoes projection in the gMLP, and \( Z \in \mathbb{R}^{T \times N \times 2d_h} \) is split into \( Z_1 \in \mathbb{R}^{T \times N \times d_h} \) and \( Z_2 \in \mathbb{R}^{T \times N \times d_h} \) in each blocks. However, unlike the single application of the Spatial Gating Unit in the gMLP, each block in Cy2Mixer employs a specialized gating mechanism designed to capture temporal, spatial, or topological dependencies, utilizing convolution or message-passing networks with different adjacency matrices in each block.

\paragraph{Temporal Block}
This block employs a \(3\times3\) convolution network to get the projection values of \(Z_2\). This is subsequently element-wise multiplied with \(Z_1\), generating the output of the Gating Unit:
\begin{align}
&\tilde{Z}_{\textit{temporal}} =  Z_{\textit{temporal},1} \odot \text{Conv}(Z_{\textit{temporal},2}),
\end{align}
where $\tilde{Z}_{\textit{temporal}} \in \mathbb{R}^{T\times N \times d_h}$. Here, 
\(Z_{\textit{temporal},1}\) and \(Z_{\textit{temporal},2} \) represent \(Z_1\) and \(Z_2\) for the Temporal block, respectively.
\paragraph{Spatial Message-Passing \& Cycle Message-Passing Blocks}
These blocks employ the MPNN for their projection function in the Gating Unit. Notably, the spatial message-passing block uses the standard adjacency matrix, \(A\), whereas the cycle message-passing block operates with the clique adjacency matrix, \(A_C\). The clique adjacency matrix, proposed by ~\cite{choi2022cycle}, represents the bases of cycles (or the first homological invariants of the graph~\citep{paton1969algorithm}) in a suitable form that enables GNNs to effectively process the desired topological features.
Following a process similar to the Temporal block, the \(\tilde{Z}\) values for both blocks are expressed as:
\begin{align}
    &\tilde{Z}_{\textit{spatial}}
    =  Z_{\textit{spatial},1} \odot \text{MPNN}(Z_{\textit{spatial},2}, A), \\
    &\tilde{Z}_{\textit{cycle}} 
    =  Z_{\textit{cycle},1} \odot \text{MPNN}(Z_{\textit{cycle},2}, A_C),
\end{align}
where $\tilde{Z}_{\textit{spatial}} \in \mathbb{R}^{T\times N \times d_h}$ and  $\tilde{Z}_{\textit{cycle}} \in \mathbb{R}^{T\times N \times d_h}$.
The final output of each block, denoted as \( Y \), is calculated by applying a linear projection to the output of Gating Unit \( \tilde{Z} \), similar to the process in Section 2.2. Considering the effective performance outcomes of incorporating tiny attention into each block in prior research~\citep{liu2021pay}, we have adopted the same structure in our model. Then we concatenate all of three outputs, namely, \(Y_{\textit{temporal}}, Y_{\textit{cycle}}\), and \(Y_{\textit{spatial}}\), using a feed-forward network as:
\begin{align}
Y_{\textit{out}} = \text{FC}\left(
    Y_{\textit{temporal}} || Y_{\textit{cycle}} || Y_{\textit{spatial}}
\right), \quad Y_{\textit{out}} \in \mathbb{R}^{T\times N \times d_h}.
\end{align}
\subsection{Output Layer}
After progressing through the stacked sequence of $L$ Cy2Mixer encoder layers, the output layer extracts the final predictions from the hidden state $Y_{out}$. Here, $T'$ represents the number of time steps to be predicted, and $d_o$ signifies the dimension of the output features.
\begin{align}
\hat{Y} = \text{FC}(Y_{out}), \quad Y \in \mathbb{R}^{T' \times N \times d_{o}}.
\end{align}

\begin{table*}
    \caption{Traffic flow prediction results for PEMS04, PEMS07, PEMS08, NYTaxi, TDrive, and CHBike. All prediction results in \textcolor{gray}{gray} are cited from available results obtained from pre-existing publications\textcolor{blue}{~\citep{liu2023spatio}}. Highlighted are the top \textbf{first} and \underline{second} results.}
    \label{table:flow_results}
    \centering
    \begin{adjustbox}{width=1.0\textwidth}
    \begin{tabular}
        {p{0.2cm}p{0.6cm}c|cccccc|ca}
        \hline
        \multicolumn{2}{c}{Datasets}   & Metric & \texttt{\textcolor{gray}{GWNET}} & \texttt{\textcolor{gray}{STGNCDE}} & \texttt{\textcolor{gray}{GMAN}} & \texttt{\textcolor{gray}{ASTGNN}} & \texttt{\textcolor{gray}{PDFormer}} & \texttt{\textcolor{gray}{STAEFormer}} & \texttt{w/o Cycle block} 
        &\texttt{Cy2Mixer}
        \\ \hline
        \multirow{3}{*}{PEMS04} & \multirow{3}{*} & MAE  & 19.36 & 19.21 & 19.14 & 18.60 & 18.36 & \underline{18.22} & 18.81 & \textbf{18.14} \\
        & & RMSE  & 31.72 & 31.09 & 31.60 & 31.03 & \underline{30.03} & 30.18 & 30.65 & \textbf{30.02} \\
        & & MAPE  & 13.30\% & 12.77\% & 13.19\% & 12.63\% & 12.00\% & \underline{11.98\%} & 12.86\% & \textbf{11.93\%} \\
        \hline
        \multirow{3}{*}{PEMS07} & \multirow{3}{*} & MAE  & 21.22 & 20.62 & 20.97 & 20.62 & 19.97 & \textbf{19.14} & 19.51 & \underline{19.45} \\
        & & RMSE  & 34.12 & 34.04 & 34.10 & 34.02 & 32.95 & \textbf{32.60} & 33.02 & \underline{32.89} \\
        & & MAPE  & 9.08\% & 8.86\% & 9.05\% & 8.86\% & 8.55\% & \textbf{8.01\%} & 8.16\% & \underline{8.11\%} \\
        \hline
        \multirow{3}{*}{PEMS08} & \multirow{3}{*} & MAE  & 15.06 & 15.46 & 15.31 & 14.97 & 13.58 & \textbf{13.46} & 13.71 & \underline{13.53} \\
        & & RMSE  & 24.86 & 24.81 & 24.92 & 24.71 & 23.41 & \underline{23.25} & 23.63 & \textbf{23.22} \\
        & & MAPE  & 9.51\% & 9.92\% & 10.13\% & 9.49\% & 9.05\% & \underline{8.88\%} & 9.01\% & \textbf{8.86\%} \\
        \hline
        \multirow{3}{*}{NYTaxi} & \multirow{3}{*} & MAE  & 13.30 & 13.28 & 13.27 & 12.98 & \textbf{12.36} & 12.61 & 12.61 & \underline{12.59} \\
        & & RMSE  & 21.71 & 21.68 & 21.66 & 21.19 & \textbf{20.18} & 20.53 & 20.53 & \underline{20.45} \\
        & & MAPE  & 13.94\% & 13.93\% & 13.89\% & 13.65\% & \textbf{12.79\%} & \underline{12.96\%} & 13.06\% & 13.03\% \\
        \hline
        \multirow{3}{*}{TDrive} & \multirow{3}{*} & MAE  & 19.55 & 19.29 & 19.10 & 18.79 & 17.79 & \textbf{16.97} & 17.48 & \underline{16.99} \\
        & & RMSE  & 36.18 & 36.12 & 36.05 & 33.93 & 31.55 & \underline{31.02} & 31.31 & \textbf{30.82} \\
        & & MAPE  & 16.56\% & 16.50\% & 16.45\% & 15.84\% & 14.68\% & \underline{13.81\%} & 13.95\% & \textbf{13.56\%} \\
        \hline
        \multirow{3}{*}{CHBike} & \multirow{3}{*} & MAE  & 4.13 & 4.11 & 4.10 & 4.02 & 3.89 & 4.03 & \underline{3.89} & \textbf{3.80} \\
        & & RMSE  & 5.81 & 5.80 & 5.79 & 5.71 & 5.48 & 5.70 & \underline{5.46} & \textbf{5.37} \\
        & & MAPE  & 30.92\% & 30.87\% & 30.91\% & 30.91\% & \underline{30.06\%} & 31.49\% & 30.10\% & \textbf{29.20\%} \\
        \hline
        \end{tabular}
    \end{adjustbox}
\end{table*}

\section{Experiments}
\paragraph{Dataset}
We evaluated Cy2Mixer on six public traffic datasets: PEMS04, PEMS07, and PEMS08, which contain only traffic data, and NYTaxi, CHBike, and TDrive, which include inflow and outflow data. For PEMS datasets, the model uses the previous hour’s data (12 time steps) to predict the next hour, while for NYTaxi, CHBike, and TDrive, it uses six steps to predict the next step. Additional details are in Appendix A.2.

\paragraph{Baseline models}
In this study, we evaluate our proposed approach against various established baseline methods in traffic forecasting.
We include GNN-based methods such as GWNet~\citep{wu2019graph} and STGNCDE~\cite{choi2022graph}.
We also include  GMAN~\citep{zheng2020gman}, ASTGNN~\cite{guo2021learning}, PDFormer~\citep{jiang2023pdformer} and STAEFormer~\citep{liu2023spatio}, all of which are self-attention-based models designed for the same task as ours.

\paragraph{Experimental settings}
We configured our experiments following standard settings for fair comparisons, with data splits of 6:2:2 for PEMS datasets and 7:1:2 for NYTaxi, TDrive, and CHBike. Conducted on an NVIDIA A100 GPU (80GB), with Python 3.10.4 and PyTorch, we used a hyperparameter search to optimize the model on validation data. Evaluation metrics included MAE, RMSE, and MAPE, averaged over 12 forecasted time steps, as detailed in Appendix A.2.

\paragraph{Results from benchmark datasets}
The comparison results between our proposed method and various baselines on traffic datasets can be found in Table \ref{table:flow_results}.
Cy2Mixer outperforms the baseline models across the majority of datasets. Comparing the results in PEMS04, it can be observed that Cy2Mixer demonstrates notable improvements, with MAE decreasing from 18.22 to 18.14, RMSE from 30.18 to 30.02, and MAPE from 11.98\% to 11.93\%. In the case of the PEMS07 dataset, the proposed model shows lower performance than STAEFormer in contrast to other datasets. This is because there are no cyclic subgraphs for the PEMS07 dataset, as observed in the Appendix A.1. This outcome underscores the substantial impact of the cycle message-passing block on the model while demonstrating that the proposed model can compete favorably with other models even without utilizing $A_{C}$ to cycle message-passing block.
The robust performance observed in the results of other datasets also demonstrates that Cy2Mixer performs well regardless of the number of predicted time steps or the number of features.

\begin{table}[t]
\caption{Ablation study on effect of cycle message-passing block for PEMS04, PEMS07, and PEMS08. Note that w/ stands for with and w/o stands for without. Time (s) refers to the preprocessing time taken to create each matrix.}
\label{tab:cycle_effect}
\centering
\renewcommand\arraystretch{1.0}
\tabcolsep=0.7mm
\resizebox{1.0\columnwidth}{!}{
\begin{tabular}{ccccc|cccc|cccc}
    
\hline
Dataset& \multicolumn{4}{c|}{PEMS04} & \multicolumn{4}{c|}{PEMS07} & \multicolumn{4}{c}{PEMS08} \\ \hline
Metric & MAE & RMSE & MAPE & Time (s) & MAE & RMSE & MAPE & Time (s) & MAE & RMSE & MAPE & Time (s)\\ \hline
w/o Cycle block & 18.81 & 30.65 & 12.86\% & - & 19.74 & 33.46 & 8.19\% & - & 13.56 & 23.45 & 8.97\% & - \\
w/ DTW & 18.44 & 30.66 & 12.16\% & 67.9732 & 19.72 & 33.35 & 8.31\% & 562.4729 & 13.65 & 23.50 & 8.94\% & 20.9622 \\ 
w/ RWSE & 18.29 & 30.00 & 11.98\% & 0.1839 & 19.57 & 34.35 & 8.07\% & 11.3208 & 13.68 & 24.00 & 8.95\% &0.0187 \\ 
w/ LapPE & 18.22 & 29.97 & 11.91\% & 0.1573 & 21.45 & 35.25 & 11.08\% & 0.9720 & 13.57 & 23.65 & 8.91\% & 0.0045 \\ 

\hline
\textbf{Cy2Mixer} & \textbf{18.14} & \textbf{30.02} & \textbf{11.93\%} & \textbf{0.0006} & \textbf{19.50} & \textbf{33.28} & \textbf{8.19\%} & \textbf{0.0281} & \textbf{13.53} & \textbf{23.22} & \textbf{8.86\%} & \textbf{0.0006}  \\ \hline
\end{tabular}
}
\end{table}
\begin{table}[t]
\caption{Case study on the structure of Cy2Mixer for PeMS04. Temporal Block, Spatial Block, and Cycle Block refer to the temporal message-passing block, spatial message-passing block, and cycle message-passing block, respectively.}
\label{tab:case_study}
\centering
\renewcommand\arraystretch{1.0}
\begin{adjustbox}{width=1.0\columnwidth}
\begin{tabular}{ccccccccc}
    
    \hline
    \multicolumn{6}{c}{Cy2Mixer (PEMS04)} & \multirow{2}{*}{MAE} & \multirow{2}{*}{RMSE} & \multirow{2}{*}{MAPE} \\ 
    \cline{1-6}
    Temporal Block & Spatial Block & Cycle Block & $A$ & $A_{C}$ & Tiny attention & & & \\ \hline
    \checkmark &   &   &   &   & \checkmark & 18.55 & 30.31 & 12.20\% \\
       & \checkmark &   & \checkmark &   & \checkmark & 22.53 & 36.33 & 15.36\% \\
    \checkmark & \checkmark  &   & \checkmark & & \checkmark & 18.81          & 30.65          & 12.86\% \\
    \checkmark &   & \checkmark &    & \checkmark & \checkmark & 18.85 & 30.68 & 12.72\% \\
       & \checkmark & \checkmark & \checkmark & \checkmark & \checkmark & 18.56 & 30.33 & 12.32\% \\
    \checkmark & \checkmark  & \checkmark  & \checkmark  & & \checkmark & 18.29 & 30.05 & 11.95\% \\
    \checkmark & \checkmark & \checkmark & \checkmark & \checkmark &   & 18.36 & 30.15 & 12.05\% \\ \hline
     \checkmark & \checkmark  & \checkmark  & \checkmark  & \checkmark  & \checkmark  & \textbf{18.14} & \textbf{30.02} & \textbf{11.93\%} \\ \hline
\end{tabular}
\end{adjustbox}
\end{table}

\section{Ablation study}

\paragraph{Effectiveness of cycle message-passing block}

We conducted an ablation study to evaluate the effect of the cycle message-passing block by testing five model variations: Cy2Mixer with the clique adjacency matrix ($A_C$), with the DTW matrix from PDFormer~\citep{jiang2023pdformer} for long-range dependencies, with alternative structural encodings, including Random-Walk Structural Encoding (RWSE)~\citep{dwivedi2021graph} and Laplacian Eigenvectors Encoding (LapPE)~\citep{rampavsek2022recipe}, and without the cycle block. Using datasets PEMS04, PEMS07, and PEMS08, results indicated that Cy2Mixer achieved the best performance when the cycle block included $A_C$. Additionally, generating $A_C$ is significantly faster than other methods like the DTW matrix, highlighting Cy2Mixer’s efficiency for spatio-temporal tasks.

\paragraph{Case study}

We conducted a case study on PEMS04 to evaluate each component’s impact in Cy2Mixer by testing variations with only the temporal or spatial message-passing blocks, and comparing models with or without the cycle message-passing block and the tiny attention mechanism. Results in Table~\ref{tab:case_study} show that the combined use of adjacency matrix $A$ and clique adjacency matrix $A_C$ in the spatial and cycle message-passing blocks significantly improves performance, as evidenced by reductions in MAE, RMSE, and MAPE. Additionally, using $A_C$ rather than $A$ in the cycle block further validated its effectiveness in enhancing Cy2Mixer’s predictive accuracy. We compared the model's performance using $A$ instead of $A_{C}$ in the cycle message-passing block to assess if the performance improvement was due to increased parameters or the influence of $A_{C}$, and this comparison demonstrated the effectiveness of $A_{C}$ in Cy2Mixer, with visualized results provided in the Appendix A.4.

\begin{table}[t]
\caption{Air pollution prediction results for \textit{KnowAir} datasets. All prediction results in \textcolor{gray}{gray} are cited from available results obtained from pre-existing publications. Highlighted are the top \textbf{first} and \underline{second} results.}
\label{tab:air_pollution_results}
\centering
\renewcommand\arraystretch{1.2}
\tabcolsep=0.7mm
\resizebox{1.0\columnwidth}{!}{
\begin{tabular}{ccc|cc|cccc}
    
\hline
Dataset& \multicolumn{2}{c|}{Sub-dataset 1} & \multicolumn{2}{c|}{Sub-dataset 2} & \multicolumn{2}{c}{Sub-dataset 3} \\ \hline
Metric & RMSE$\downarrow$ & CSI({\%})$\uparrow$ & RMSE$\downarrow$ & CSI({\%})$\uparrow$ & RMSE$\downarrow$ & CSI({\%})$\uparrow$ \\ \hline
\texttt{\textcolor{gray}{GRU}} & 21.00 ± 0.17 & 45.38 ± 0.52 & 32.59 ± 0.16 & 51.07 ± 0.81 & 45.25 ± 0.85 & 59.40 ± 0.01 \\
\texttt{\textcolor{gray}{GC-LSTM}} & 20.84 ± 0.11 & 45.83 ± 0.43 & 32.10 ± 0.29 & 51.24 ± 0.13 & 45.01 ± 0.81 & 60.58 ± 0.14 \\
\texttt{\textcolor{gray}{$\text{PM}_{2.5}$-GNN}} & 19.93 ± 0.11 & \underline{48.52 ± 0.48} & 31.37 ± 0.34 & \textbf{52.33 ± 1.06} & 43.29 ± 0.79 & \underline{61.91 ± 0.78} \\
\texttt{\textcolor{gray}{$\text{PM}_{2.5}$-GNN no PBL}} & 20.46 ± 0.18 & 47.43 ± 0.37 & 32.44 ± 0.36 & 51.05 ± 1.15 & 44.71 ± 1.02 & 60.64 ± 0.84 \\
\texttt{\textcolor{gray}{$\text{PM}_{2.5}$-GNN no export}} & 20.54 ± 0.16 & 45.73 ± 0.58 & 31.91 ± 0.32 & 51.54 ± 1.27 & 43.72 ± 1.03 & 61.52 ± 0.95\\
\hline
\texttt{w/o Cycle block} & \underline{19.76 ± 0.13} & 47.95 ± 0.33 & \underline{31.31 ± 0.20} & \underline{52.03 ± 0.95} & \underline{43.27 ± 0.41} & 61.65 ± 0.52 \\
\rowcolor{Gray}\texttt{\textbf{Cy2Mixer}} & \textbf{19.34 ± 0.13} & \textbf{48.58 ± 0.56} & \textbf{31.29 ± 0.19} & 51.64 ± 0.90 & \textbf{43.19 ± 0.98} & \textbf{62.06 ± 0.69} \\
\hline
\end{tabular}
}
\end{table}

\paragraph{Additional experiments for different spatio-temporal learning tasks.}
We conducted additional experiments to validate the proposed Cy2Mixer not only for traffic forecasting but also for other various spatio-temporal tasks. Among these tasks, we specifically verified the performance of our model in the context of air pollution prediction. A whole 4-year dataset \textit{KnowAir} was used for predicting particles smaller than 2.5$\mu\mathrm{m}$ ($\text{PM}_{2.5}$) concentrations~\citep{wang2020pm2}.
The dataset covers in total 184 cities expressed as nodes, and is split into 3 sub-datasets based on dates. Since the comparison models in this context differ from the models used in the main text, Cy2Mixer was compared with models designed for air pollution prediction; GRU, GC-LSTM~\citep{qi2019hybrid}, and $\text{PM}_{2.5}$-GNN~\citep{wang2020pm2}. The results of this comparison were evaluated using root mean squre error (RMSE) and critical success index (CSI), which are commonly used meteorological metrics, are presented in Table~\ref{tab:air_pollution_results}. Cy2Mixer consistently demonstrated the highest predictive performance across most datasets and notably, it achieved this without relying on specific domain knowledge.

\section{Conclusion}
In this paper, we have mathematically investigated the effects of topological non-trivial invariants on capturing the complex dependencies of spatio-temporal graphs. Through our investigation, we gained insight into how the homotopy invariance of fundamental groups and homology groups can be a contributing factor in influencing the predictive performance of spatio-temporal graphs. We then introduce a simple yet novel model, Cy2Mixer, based on the mathematical background and inspired by gMLP. Cy2Mixer comprises of three major components: a temporal block, a spatial message-passing block, and a cycle message-passing block. Notably, the cycle message-passing block enriches topological information in each of the Cy2Mixer encoder layers, drawn from cyclic subgraphs. Indeed, Cy2Mixer achieves state-of-the-art or second-best performance on traffic forecast benchmark datasets. We further investigate the effects of the cycle message-passing block on benchmark datasets by comparing with the DTW method, which makes a new adjacency matrix based on the time-dependent similarity for each node. Compared to the DTW method, the cycle message-passing block captures the spatio-temporal dependency more effectively with a significantly lower computational cost. 

\bibliographystyle{unsrtnat}
\bibliography{log24}


\newpage
\appendix
\section{Appendix}
\subsection{Dataset description}
Detailed statistical information of six traffic datasets are outlined in Table~\ref{tab:data_description}. The first three datasets are graph-based datasets, each with a single feature, while the latter three are grid-based datasets with two features. The node counts for the grid-based datasets are 75 (15$\times$5), 270 (15$\times$18), and 1,024 (32$\times$32), respectively. \textcolor{black}{The term ``\# Cycles" indicates the number of the cycle bases of graphs, and the term ``Average Magnitude \# Cycles" denotes the average number of nodes present in a cycle subgraph of each graph.}

\begin{table}[!ht]
  \centering
  \caption{Description of statistical information of six traffic datasets.}
  \vskip 0.15in
  \resizebox{\columnwidth}{!}{
    \begin{tabular}{ccccccccc}
    
    \hline
    Datasets & \# Nodes & \# Edges & \# Timesteps & \# Time Interval & Time range & \# Features & \# Cycles & Average Magnitude \# Cycles\\
    \midrule
    PEMS04 & 307   & 340   & 16,992 & 5 min  & 01/01/2018-02/28/2018 & 1 & 45 & 4.9111 \\
    PEMS07 & 883   & 866   & 28,224 & 5 min  & 05/01/2017-08/31/2017 & 1 & 0 & 0\\
    PEMS08 & 170   & 295   & 17,856 & 5 min  & 07/01/2016-08/31/2016 & 1 & 105 & 7.8571\\
    \midrule
    NYTaxi & 75 & 484   & 17,520 & 30 min & 01/01/2014-12/31/2014 & 2 & 168 & 4.8810\\
    TDrive & 1,024 & 7,812  & 3,600  & 60 min & 02/01/2015-06/30/2015 & 2 & 714 & 7.9174\\
    CHBike & 270 & 1,966  & 4,416  & 30 min & 07/01/2020-09/30/2020 & 2 & 2883 & 14.4707\\
    
    \bottomrule
    \end{tabular}%
    }
  \label{tab:data_description}%
\end{table}%

\subsection{Hyperparameter search}

We conducted a hyperparameter search to find the optimal model, and hyperparameters for each datasets are listed in Table~\ref{tab:hyperparameter}. Note that the embedding dimensions, $d_{f}$, $d_{t}$, and $d_{a}$, followed the settings of previous research~\cite{liu2023spatio}. The search ranges were $\{16, 32, 64, 128\}$ for hidden dimension $d_{h}$, $\{2, 3, 4, 5, 6\}$ for number of layers, and $\{0, 0.2, 0.4, 0.6, 0.8\}$ for dropout rate, respectively. Considering previous research has demonstrated the effectiveness of incorporating tiny attention, we conducted experiments in our study to compare the performance when tiny attention is added and when it is not. The selection of the optimal model was based on its performance on the validation set. We performed experiments on STAEFormer~\citep{liu2023spatio} framework.

\begin{table}[!ht]
  \centering
  \caption{Hyperparameters for six traffic datasets.}
  \vskip 0.15in
  \resizebox{\columnwidth}{!}{
    \begin{tabular}{cccccccccccc}
    
    \hline
    Datasets & \# Layers & $d_{f}$ & $d_{t}$ & $d_{a}$ & $d_{h}$ & Batch size & Dropout & Weight decay & Learning rate & Learning rate decay & Tiny attention\\
    \midrule
    PEMS04 & 3   & 24   & 24 & 80  & 152 & 16 & 0.4 & 0.0005 & 0.001 & 0.1 & O\\
    PEMS07 & 4   & 24   & 24 & 80  & 152 & 16 & 0.4 & 0.001 & 0.001 & 0.1 & X\\
    PEMS08 & 3   & 24   & 24 & 80  & 152 & 16 & 0.1 & 0.0015 & 0.001 & 0.1 & O\\
    \midrule
    NYTaxi & 5   & 24   & 24 & 80  & 256 & 16 & 0.4 & 0.05 & 0.001 & 0.1 & X\\
    TDrive & 6   & 24   & 24 & 80  & 256 & 16 & 0.4 & 0.05 & 0.001 & 0.1 & X\\
    CHBike & 3   & 24   & 24 & 80  & 256 & 16 & 0.4 & 0.05 & 0.001 & 0.1 & O\\
    \bottomrule
    \end{tabular}%
    }
  \label{tab:hyperparameter}%
\end{table}%

\subsection{Comparison of the adjacency matrix, clique adjacency matrix, and DTW matrix.}

\begin{figure}[ht!]
  \centering
  \begin{subfigure}{}
    \includegraphics[width=0.31\linewidth]{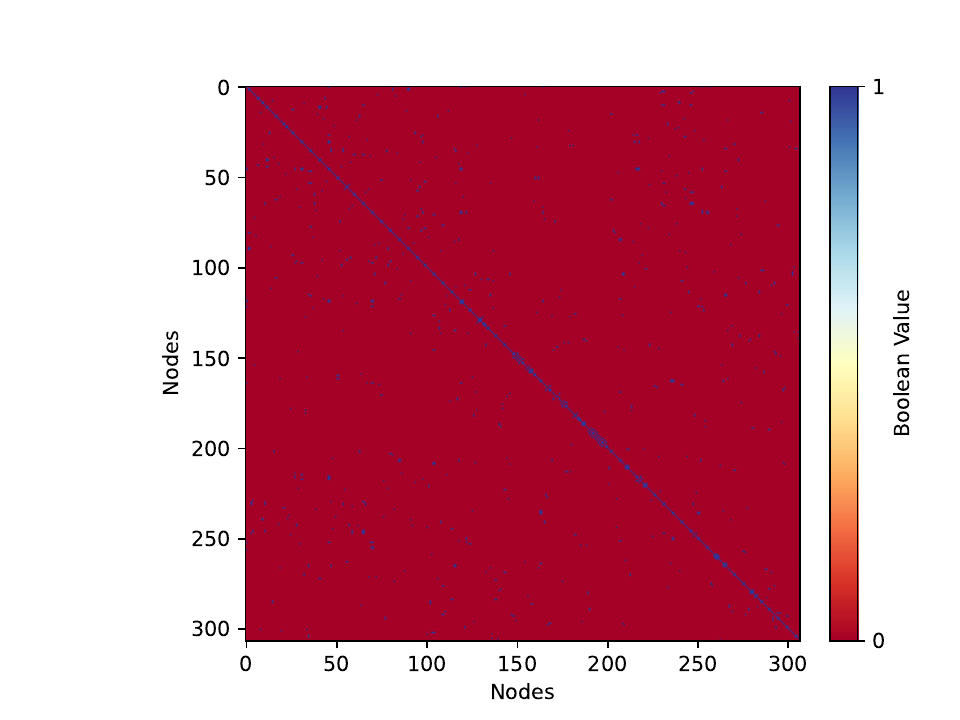}
  \end{subfigure}
  \begin{subfigure}{}
    \includegraphics[width=0.31\linewidth]{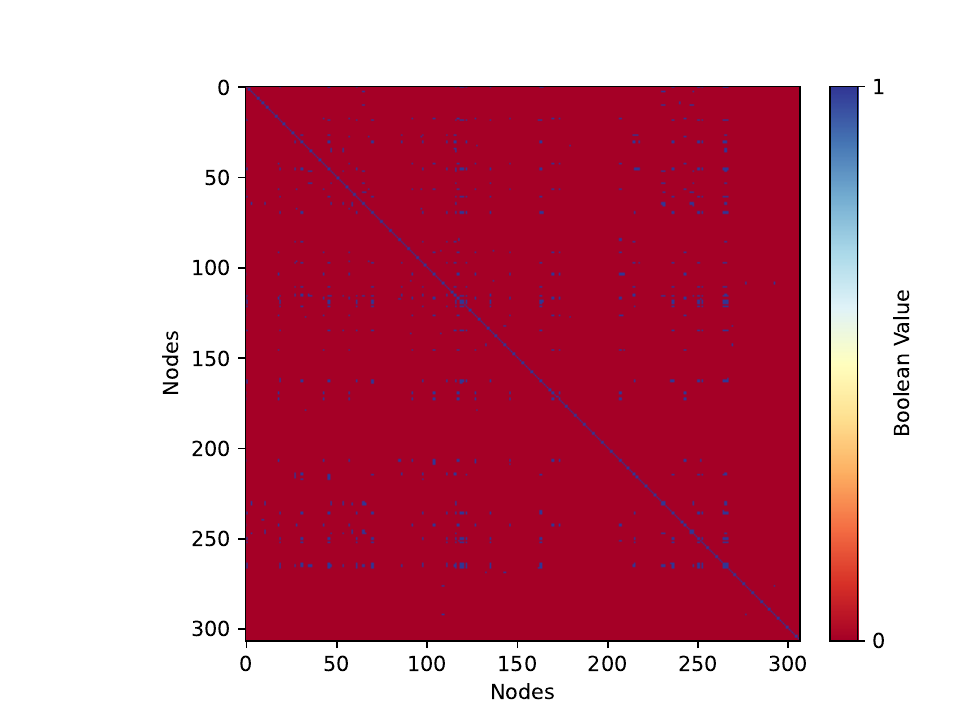}
  \end{subfigure}
  \begin{subfigure}{}
    \includegraphics[width=0.31\linewidth]{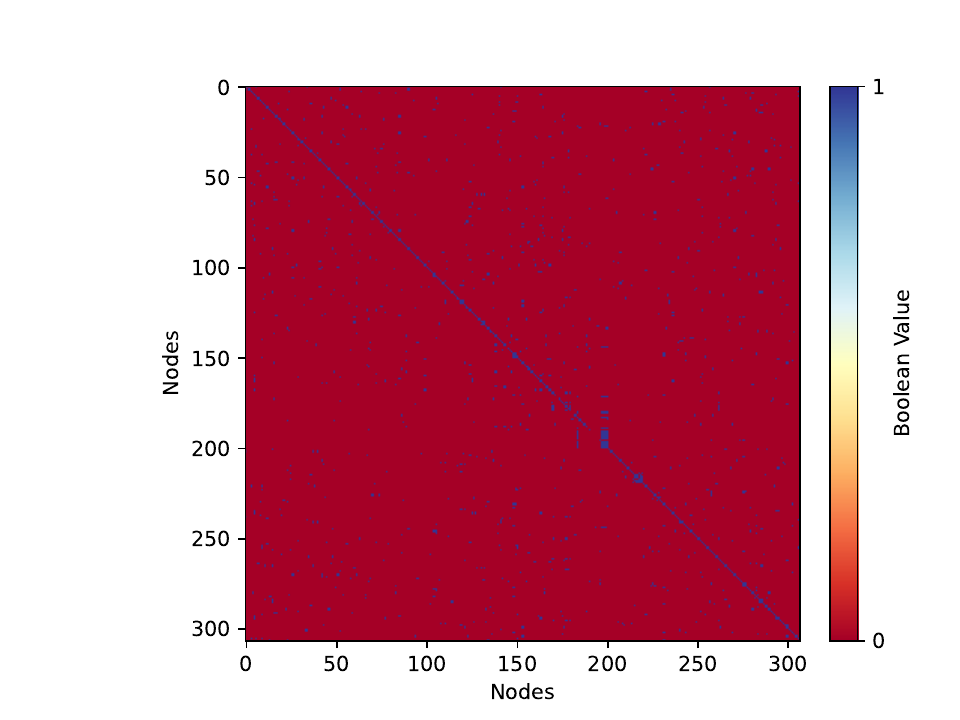}
  \end{subfigure}
  \caption{Three matrices of PEMS04:  (a) Adjacency matrix $A$, (b) Clique adjacency matrix $A_{C}$, and (c) Matrix constructed by DTW algorithm, respectively. The figure demonstrates that the clique adjacency matrix we use can capture topological information different from the standard adjacency matrix and DTW matrix.}
\label{Fig:matrices}
\end{figure}

In this section, a detailed comparison is conducted between the adjacency matrix and clique adjacency matrix used in Cy2Mixer, and the DTW matrix employed in PDFormer~\cite{jiang2023pdformer}. The three matrices constructed from the traffic network are visually demonstrated in Fig.~\ref{Fig:matrices}. DTW algorithm is a similarity measure between two time-series that allows for non-linear alignment of the time series. In the context of PDFormer, DTW is used to compute the similarity of the historical traffic flow between nodes to identify the semantic neighbors of each node by calculating a pairwise distance matrix encompassing all combinations of historical traffic flow sequences. This distance matrix enables the computation of a cumulative distance matrix, representing the minimum distance between two time-series up to a specific point. Finally, the cumulative distance matrix facilitates the determination of the optimal warping path, signifying the most favorable alignment between two time-series. DTW algorithm allows the model to capture complex traffic patterns that are not easily captured by linear alignment methods, but has significant drawbacks: it is computationally time-consuming to compute and often relies on heuristic thresholds and parameters. To overcome these limitations and capture the desired spatio-temporal dependencies, we employ the use of the clique adjacency matrix. The clique adjacency matrix can be computed more efficiently than DTW since it is not dependent on time. Furthermore, it provides the model with richer topological information within the graph. The figure demonstrates that the clique adjacency matrix we use can capture topological information different from the standard adjacency matrix and DTW matrix.

\subsection{Visualization of prediction results for each node}
\begin{figure}[ht!]
  \centering
  \begin{subfigure}{}
    \includegraphics[width=0.4\linewidth]{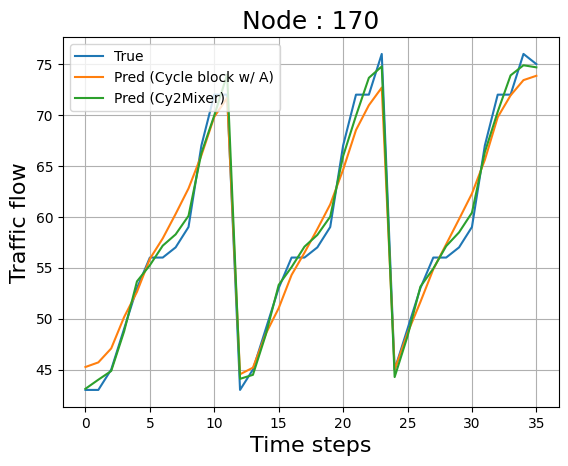}
  \end{subfigure}
  \begin{subfigure}{}
    \includegraphics[width=0.4\linewidth]{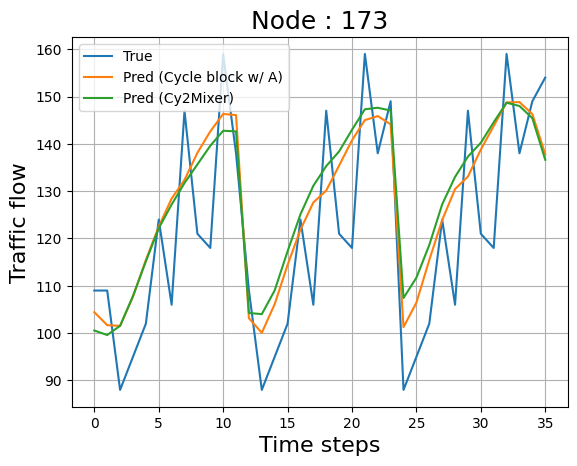}
  \end{subfigure}
  \begin{subfigure}{}
    \includegraphics[width=0.4\linewidth]{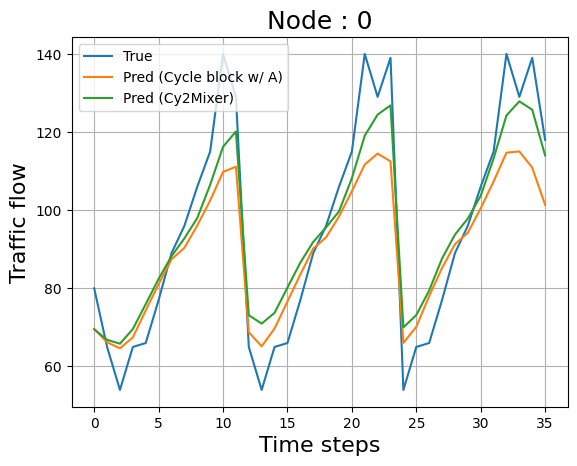}
  \end{subfigure}
  \begin{subfigure}{}
    \includegraphics[width=0.4\linewidth]{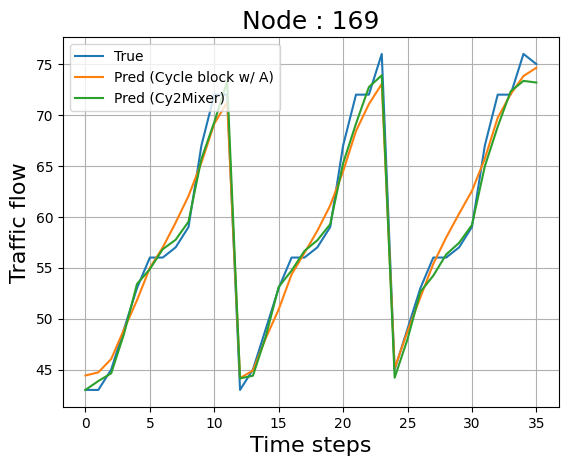}
  \end{subfigure}
  
  \caption{Visualization of prediction results for Node 170, Node 173 (connected to Node 170 in the adjacency matrix), and Nodes 0 and 169 (connected to Node 170 in the clique adjacency matrix) in the PEMS04 dataset.}
\label{Fig:prediction_result}
\end{figure}

In this section, we present the actual prediction results for each node in Cy2Mixer, as shown in Fig~\ref{Fig:prediction_result}. To determine whether the cycle block had significant impact on predictions, we compare the results of Cy2Mixer and a model that uses adjacency matrix $A$ instead of clique adjacency matrix $A_{C}$ in the cycle block. The experiment was conducted on 36 time steps, for Node 170, Node 173 connected to Node 170 through the adjacency matrix, and Node 0 and 169 connected through the clique adjacency matrix. While both models show similar performance for Node 170 and 173, Cy2Mixer demonstrates better prediction accuracy for Node 0 and 169. This indicates that the cycle block effectively incorporates information from the clique adjacency matrix.

\subsection{Comparison of the influence of the predefined adjacency matrix}
In this section, we compared whether a predefined adjacency matrix, calculated based on the distances between nodes, makes a significant difference in the model's performance. Recent studies in traffic prediction problems have explored not using a predefined adjacency matrix, instead learning it alongside the model~\cite{wu2020connecting}. To conduct this comparison, we adopted methods from previous studies to create an adjacency matrix and then generated a clique adjacency matrix from it for training in Cy2Mixer. The results of this approach are presented in Table~\ref{tab:predefined_effect}. The findings indicate that Cy2Mixer does not show significant differences in performance between using a predefined $A$ and a learned $A$. Particularly, since learning $A$ and subsequently generating $A_{C}$ is costly, this study utilized a predefined $A$.

\begin{table}[ht!]
\caption{Ablation study on effect of predefined adjacency matrix PEMS04, PEMS07, and PEMS08. Note that w/ stands for with and w/o stands for without.}
\vskip 0.15in
\label{tab:predefined_effect}
\centering
\renewcommand\arraystretch{1.2}
\tabcolsep=0.7mm
\resizebox{0.8\columnwidth}{!}{
\begin{tabular}{cccc|ccc|ccc}
    
\hline\hline
Dataset& \multicolumn{3}{c|}{PEMS04} & \multicolumn{3}{c|}{PEMS07} & \multicolumn{3}{c}{PEMS08} \\ \hline\hline
Metric & MAE & RMSE & MAPE & MAE & RMSE & MAPE & MAE & RMSE & MAPE \\ \hline\hline
w/o Cycle block & 18.81 & 30.65 & 12.86\% & 19.74 & 33.46 & 8.19\% & 13.56 & 23.45 & 8.97\% \\
w/ learned $A$, $A_{C}$ & 18.81 & 30.54 & 12.22\% & 19.74 & 33.37 & 8.33\% & 13.56 & 23.46 & 8.92\% \\ \hline\hline
\textbf{Cy2Mixer} & \textbf{18.14} & \textbf{30.02} & \textbf{11.93\%} & \textbf{19.50} & \textbf{33.28} & \textbf{8.19\%} & \textbf{13.53} & \textbf{23.22} & \textbf{8.86\%} \\ \hline\hline
\end{tabular}
}
\end{table}

\subsection{Proof of Theorem \ref{theorem:cycle_temporal}}
The theorem is a corollary of \citet{Ha02}[Proposition 1.18, Corollary 2.11]. To elaborate, the map $\pi_{t_0}: \mathcal{G} \times I \to \mathcal{G}$ is a homotopy equivalence between the space $\mathcal{G} \times I$ representing the temporal traffic data and the space $\mathcal{G}$ representing the traffic network. The map induces an isomorphism between the fundamental groups and the homology groups of two topological spaces with rational coefficients. That is, these two maps are isomorphisms of groups for all indices $i \geq 0$:
    \begin{align*}
    \begin{split}
        (\pi_{t_0})_*: \pi_1(\mathcal{G} \times I) \to \pi_1(\mathcal{G}) \\
        (\pi_{t_0})_*: H^i(\mathcal{G} \times I, \mathbb{Q}) \to H^i(\mathcal{G}, \mathbb{Q})
    \end{split}
    \end{align*}
    The isomorphisms between the fundamental groups $\pi_1$ and the first homology groups $H_1$ indicate that the set $\pi_{t_0}(\mathcal{C}_{\mathcal{G} \times I})$ has to be a cycle basis of $\mathcal{G}$.

\paragraph{Mathematical relevance of Theorem \ref{theorem:cycle_temporal} to temporal traffic data}
{\textcolor{black} Theorem \ref{theorem:cycle_temporal} shows in particular that any cyclic subgraph of the traffic signal $\mathcal{G} \times I$ is comprised of nodes of form $(c_1, t_1), (c_2,t_2), \cdots, (c_n, t_n) \in \mathcal{G} \times I$, where the nodes $c_1, c_2, \cdots, c_n$ are on a common cyclic subgraph of $\mathcal{G}$, and the coordinates $t_i$ correspond to different temporal instances.}
Theorem \ref{theorem:cycle_temporal} mathematically demonstrates that the topological non-trivial invariants of a traffic network $\mathcal{G}$ can become a contributing factor for influencing the temporal variations measured among the nodes of traffic dataset, a potential aspect of temporal traffic data that may not be fully addressed from solely analyzing the set of pre-existing edges connecting the nodes of a traffic network. {\textcolor{black} To elaborate, topological non-trivial invariants of $\mathcal{G}$ elucidate restrictions associated to constructing a global traffic signal $X: \mathcal{G} \times I \to \mathbb{R}^{N \times C}$ from coherently gluing a series of local temporal traffic signals $\{X_t\}$ measured at each time $t$. This originates from previously studied mathematical insights that obstructions in gluing continuous functions $f_\alpha: U_\alpha \to \mathbb{R}^k$ defined over collections of open covers $U_\alpha \subset Y$ of a topological space (such that $\cup_\alpha U_\alpha = Y$) to a continuous function $f: Y \to \mathbb{R}^k$ can be detected from topological (or cohomological) invariants of the topological space $Y$ (see for example Chapter 2 or 3 of \cite{Ha77}). Given that traffic forecasting beyond time $t = t_e$ requires a thorough understanding of traffic signals $X: \mathcal{G} \times [0, t_e] \to \mathbb{R}^{N \times C}$, we can hence conclude that a potential candidate to boost performances of traffic forecasting algorithms is to effectively incorporate topological invariants of traffic networks $\mathcal{G}$.}

\paragraph{Exemplary Illustration} {We provide an exemplary illustration on how cycle structures are closely relevant to determining the range of traffic data (or traffic signals) that can be measured on a given traffic network.}

\begin{figure}
    \centering
    \includegraphics[width=\linewidth]{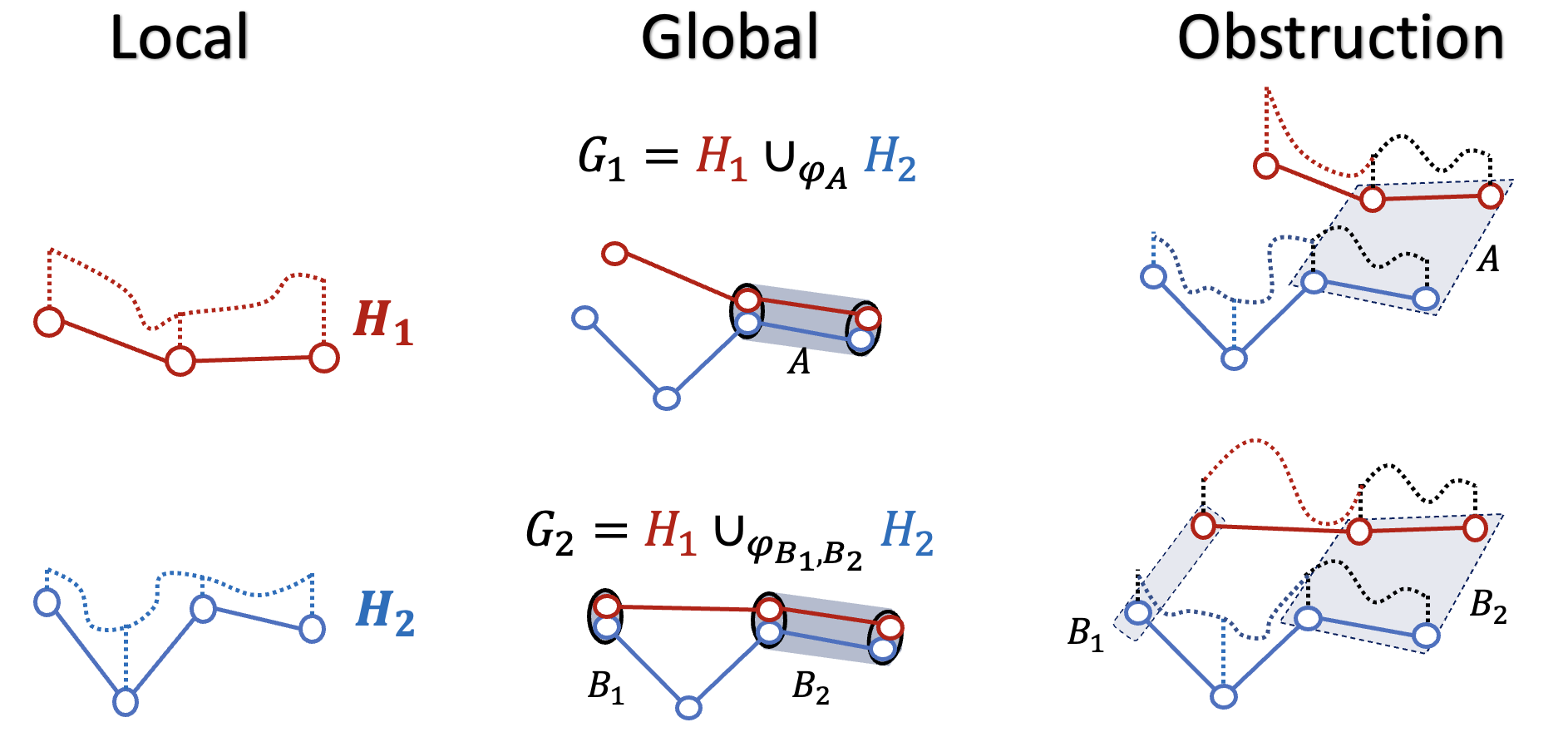}
    \caption{An exemplary illustration of obstructions cycle structures may impose on the range of traffic data or signals that can be measured in a given traffic network.}
    \label{fig:cycle-obstruction}
\end{figure}

As shown in Figure \ref{fig:cycle-obstruction}, suppose we are given with two local traffic networks (graphs) without cycles $H_1$ (the red graph) and $H_2$ (the blue graph). The dotted lines above the graphs represent an example of a traffic signal that can be measured over each local traffic network. Let $G_1$ and $G_2$ be two global traffic networks (graphs) which can be obtained from gluing the two local traffic networks in a different manner. The traffic network $G_1$ glues two nodes and one edge of $H_1$ and $H_2$ together (along the subnetwork notated as $A$), which does not add new cyclic subgraph to $G_1$. On the other hand, the traffic network $G_2$ glues three nodes and one edge of $H_1$ and $H_2$ together (along the subnetworks notated as $B_1$ and $B_2$), which adds a new cyclic subgraph to $G_2$. By gluing the two local traffic networks together, we introduce a new obstruction to traffic signals that can be measured over $G_1$ and $G_2$. The signals measured over subnetworks of local traffic network $H_1$ and $H_2$ must be identical to one another. Figure \ref{fig:cycle-obstruction} clearly shows that $G_2$ has more obstructions on the traffic signal that can be measured in comparison to $G_1$. To elaborate, any traffic signal that can be measured on $G_2$ must satisfy the identical signals over the subnetworks $B_1$ and $B_2$, whereas $G_1$ only requires to satisfy the identical signals over the subnetworks $A$, which is isomorphic to the subnetwork $B_2$. This illustration hints that demonstrates that determining cycle structures inherent in a traffic network is crucial to understanding additional obstructions imposed over traffic signals that can be measured or predicted in the given network. Because conventional graph neural networks have limited capabilities in discerning cyclic substructures of graph datasets (as shown in \cite{choi2022cycle}), adding additional architectural component which can detect cycle structures of traffic networks to the algorithm can be of crucial importance in improving accuracies in forecasting traffic signals over any traffic networks.

\subsection{Clique adjacency matrix} \; \; The motivation for using the clique adjacency matrix in Cy2Mixer stems from recent advances in distinguishing graph structures through the theory of covering spaces~\citep{choi2022cycle}. Traditional GNNs represent graphs $G$ and $H$ as identical if their universal covers—essentially infinite graphs created by unfolding the original graphs at each node—are isomorphic, and their node features match upon this unfolding. However, these universal covers inherently lack cyclic subgraphs, which are crucial structures in the original graphs. To improve the ability of GNNs to discern these cyclic features, the clique adjacency matrix ($A_C$) is used in Cy2Mixer. This matrix effectively transforms the geometry of the universal covers, incorporating cyclic subgraph information and thereby enhancing the model’s ability to capture and leverage topological nuances within the graph structure. The clique adjacency matrix $A_C$ is generated as follows: given an undirected graph $G := (V, E)$, we first establish a cycle basis $B_G$ of $G$, representing a set of cyclic subgraphs that form a basis for the cycle space (or the first homology group) of $G$. The clique adjacency matrix $A_C$ is then constructed as the adjacency matrix of the union of complete subgraphs formed by each cycle in $B_G$. This is achieved by adding all possible edges among the nodes within each cycle basis element $B \in B_G$, enabling $A_C$ to represent cyclic structures comprehensively in Cy2Mixer.

\begin{table*}
    \caption{Traffic flow prediction results for METR-LA. Highlighted are the top \textbf{first} and \underline{second} results.}
    \label{table:flow_results_metrla}
    \centering
    \begin{adjustbox}{width=1.0\textwidth}
    \begin{tabular}{p{0.2cm}p{1.2cm}c|cccccccc}
    \hline
    \multicolumn{2}{c}{Dataset}   & Metric & \texttt{\textcolor{gray}{STGNCDE}} & \texttt{\textcolor{gray}{GMAN}} & \texttt{\textcolor{gray}{PDFormer}} & \texttt{\textcolor{gray}{HGCN}} & \texttt{\textcolor{gray}{HIEST}} & \texttt{\textcolor{gray}{MTGNN}} & \texttt{\textcolor{gray}{STAEFormer}} & \texttt{Cy2Mixer} \\
    \hline
    \multirow{9}{*}{\rotatebox{90}{METR-LA}} & \multirow{3}{*}{Horizon 3} & MAE  & 3.35 & 3.53 & 2.83 & 2.80 & 2.71 & \textbf{2.66} & \underline{2.68} & 2.70 \\
    & & RMSE  & 7.10 & 6.12 & 5.55 & 5.40 & \underline{5.20} & \textbf{5.12} & 5.22 & 5.30 \\
    & & MAPE  & 8.07\% & 10.36\% & 7.46\% & 7.38\% & \underline{6.97\%} & \textbf{6.83\%} & 7.01\% & 7.13\% \\
    \cline{2-11}
    & \multirow{3}{*}{Horizon 6} & MAE  & 4.00 & 3.95 & 3.23 & 3.23 & 3.13 & \underline{3.03} & \textbf{3.00} & \underline{3.03} \\
    & & RMSE  & 8.59 & 6.93 & 6.64 & 6.48 & 6.23 & \textbf{6.09} & \underline{6.12} & 6.22 \\
    & & MAPE  & 9.86\% & 11.79\% & 9.16\% & 9.06\% & 8.62\% & \textbf{8.26\%} & \underline{8.34\%} & 8.48\% \\
    \cline{2-11}
    & \multirow{3}{*}{Horizon 12} & MAE  & 4.84 & 4.46 & 3.64 & 3.68 & 3.59 & 3.45 & \textbf{3.37} & \underline{3.39} \\
    & & RMSE  & 10.52 & 7.81 & 7.67 & 7.53 & 7.37 & \underline{7.15} & \textbf{7.09} & 7.21 \\
    & & MAPE  & 12.34\% & 13.55\% & 10.87\% & 10.84\% & 10.42\% & 9.99\% & \textbf{9.94\%} & \underline{9.96\%} \\
    \hline
    \end{tabular}
    \end{adjustbox}
\end{table*}

\subsection{Performance Analysis of Cy2Mixer on the METR-LA Dataset}
We also performed an experiment in the METR-LA dataset~\citep{chen2001freeway}, which is a traffic speed dataset that records the average vehicle speeds across various sensors in Los Angeles. In our experiments on METR-LA, we evaluated the model’s performance at different forecasting horizons, specifically for time steps at horizons of 3, 6, and 12, which correspond to 15, 30, and 60 minutes into the future. The data was split into training, validation, and testing sets with ratios of 7:1:2, respectively. Table \ref{table:flow_results_metrla} presents the comparison results between our proposed method, Cy2Mixer, and various baseline models on the METR-LA dataset. Cy2Mixer demonstrates competitive performance across different forecasting horizons when compared to existing models. While Cy2Mixer does not consistently outperform all baseline models on the METR-LA dataset, it delivers results that are on par with the leading methods. The slight differences in performance metrics could be attributed to the specific characteristics of the METR-LA dataset, including its network structure and traffic patterns. Additionally, Cy2Mixer shows consistent performance across Horizons 3, 6, and 12, unlike MTGNN~\citep{wu2020connecting} and HIEST~\citep{ma2023rethinking}, whose performance deteriorates at Horizon 12.

\end{document}